\documentclass[journal]{IEEEtran}
\usepackage{graphicx}
\usepackage{subfigure}
\usepackage{multirow}
\usepackage{array}
\usepackage{footnote}

\usepackage[ruled,linesnumbered]{algorithm2e}

\usepackage{dsfont}
\usepackage{amsfonts,amsmath}
\usepackage{url,cite}
\usepackage{color}

\def\cred{\textcolor{black}}
\def\cblue{\textcolor{black}}
\begin{document}

\title{Hyperspectral Unmixing via Nonnegative Matrix Factorization with Handcrafted and Learnt Priors}

\author{Min~Zhao,~\IEEEmembership{Student Member,~IEEE,}
          Tiande~Gao,
        Jie~Chen,~\IEEEmembership{Senior Member,~IEEE,}\\
        Wei~Chen,~\IEEEmembership{Senior Member,~IEEE} \vspace{-5mm}
\thanks{M. Zhao, T. Gao, J. Chen are with School of Marine Science and \cred{Technology, Northwestern} Polytechnical University, China. W. Chen is with State Key Laboratory of Rail Traffic Control and Safety, Beijing Jiaotong University, China. (Corresponding author: J. Chen, dr.jie.chen@ieee.org).}
}

\markboth{IEEE xx,~Vol.~xx, No.~xx,~2020}%
{Shell \MakeLowercase{\textit{et al.}}: Hyperspectral Unmixing via Nonnegative Matrix Factorization with Handcrafted and Learnt Priors}

\maketitle

\begin{abstract}
Nowadays, nonnegative matrix factorization (NMF) based methods have been
widely applied to blind spectral unmixing. Introducing proper regularizers
to NMF is crucial for mathematically constraining the solutions  and
physically exploiting  spectral and spatial properties of images.
Generally, properly handcrafting regularizers and solving the associated
complex optimization problem are non-trivial tasks. In our work, we
propose an NMF based unmixing framework which jointly uses  a handcrafting
regularizer and a learnt regularizer from data. \cblue{we plug learnt
priors of abundances where the associated subproblem can be addressed
using various image denoisers, and we consider an $\ell_{2,1}$-norm
regularizer to the abundance matrix to promote sparse unmixing results.}
The proposed framework is flexible and extendable.  Both synthetic data
and real airborne data are conducted to confirm the effectiveness of our
method.
\end{abstract}


\IEEEpeerreviewmaketitle
\section{Introduction}
W{ith} the development of remote sensing and hyperspectral sensors,
hyperspectral imaging has been widely adopted in many fields. However, due
to the low spatial resolution of hyperspectral sensors and the complexity of
material mixing process, an observed pixel may contain several different
materials. Hyperspectral unmixing analyzes the mixed pixel at a subpixel
level, by decomposing the mixed pixel into a set of endmembers and their
corresponding fractional
abundances~\cite{keshava2002spectral,bioucas2012hyperspectral}.

In the past few decades, many hyperspectral unmixing methods have been
proposed, including step-by-step unmixing methods (first extracting
endmembers and then estimating abundances) and simultaneous decomposing
methods (simultaneously determining endmembers and abundances ). Nonnegative
matrix factorization (NMF) is a popular simultaneously decomposing method.
It represents a hyperspectral image by a product of two nonnegative matrices
under the linear mixing assumption, namely, one represents the endmember
matrix, and the other represents the abundance matrix~\cite{li2016robust}.
However, NMF is a nonconvex problem that does not admit a unique solution.
To overcome this drawback,  regularizers are proposed to be added to the
objective function to mathematically constrain  the space of solutions as
well as physically exploit the spatial and spectral properties of
hyperspectral images. So far, sparsity and spatial smoothness are the most
useful and common regularizers~\cite{lu2013double,salehani2017smooth}. As
mixed pixels often contain a subset of endmembers, sparsity-promoting
regularizers have been widely applied in NMF based unmixing
methods~\cite{wang2016hypergraph,lu2019subspace}. The
$\ell_1$-norm~\cite{he2017total}, and more generally, $\ell_p$-norm
($0<p<1$)~\cite{sigurdsson2016blind,salehani2017smooth}  regularizers are
introduced to produce sparse unmixing results. Thus, many NMF works use the
$\ell_p$-norm to enhance the sparsity of results. Spatial smoothness is an
inherent characteristic of natural hyperspectral images. Many works
integrate the total-variation (TV) regularization to enhance the
similarities of neighbor
pixels~\cite{feng2018hyperspectral,sigurdsson2016blind}. Reweight TV
regularization is also proposed to adaptively capture the smooth structure
of abundance maps~\cite{he2017total,feng2019hyperspectral}. Graph based
regularization is also a sophisticated  choice to cope with spatial
information, especially images with complex structures. Many works, such as
MLNMF~\cite{shu2015multilayer}, AGMLNMF~\cite{tong2020adaptive} and
SGSNMF~\cite{wang2017spatial}, use graph regularization to capture the
comprehensive spatial information. SSRNMF~\cite{huang2019spectral} uses the
$\ell_{2,1}$-norm and the $\ell_{1,2}$-norm to jointly capture spatial and
spectral priors. \cred{However, all these methods use handcrafted
regularizers to learn the inherent information of hyperspectral data and
enhance the unmixing accuracy of NMF.} \cred{Generally, designing a proper
regularization is a non-trivial task}, and complex regularizers may increase
the difficulty of solving the optimization problem.  Several plug-and-play
methods have been proposed to solve various hyperspectral image inverse
problems~\cite{sreehari2016plug,teodoro2018convergent,wang2020learning}.
However\cred{,} this strategy has not been considered in the hyperspectral
unmixing with NMF.

In this letter, we propose \cred{a novel} NMF based unmixing framework which
jointly considers both handcrafted and learnt regularizers  to fully
investigate the prior information embedded in hyperspectral images. To be
specific, \cred{we plug the learnt spectral and spatial information by using
a denoising operator. Using such an operator gets rid of manually designing
regularizer and solving the complex optimization problems. A variety of
denoisers can be plugged into the framework, which make the method
flexible and extendable.} \cred{Further, handcrafted prior is also kept to encode some physically explicit
priors.} \cred{We add an $\ell_{2,1}$-norm to the objective
function to enhance the sparsity of abundances as an example, as the number of endmembers
is usually larger than the number of materials existing in a mixed pixel.}

\section{Problem Formulation}
We consider the linear mixture model (LMM), which assumes that an observed
pixel is a linear combination of a set of endmembers and their associated
fractional abundances. Let $\mathbf{R}\in\mathbb{R}^{L\times N}$ be an
observed hyperspectral image, where $L$ is the number of spectral bands, $N$
is the number of pixels. The LMM for a hyperspectral image can be expressed
as:
\begin{equation}\label{eq.LMM}
  \mathbf{R} = \mathbf{EA}+\mathbf{N},
\end{equation}
where $\mathbf{E}$ is an $L\times P$ matrix, which denotes the endmember
spectra library with $P$ the number of endmembers, and $\mathbf{A}$ is a
$P\times N$ matrix representing the abundance matrix, and $\mathbf{N}$
denotes an i.i.d. zero-mean Gaussian noise matrix. For the physical
characteristics of hyperspectral data, the endmember matrix $\mathbf{E}$ is
required to satisfy the endmember nonnegative constraint (ENC), and the
abundance matrix $\mathbf{A}$ is required to satisfy the abundance
nonnegative constraint (ANC) and sum-to-one constraint (ASC), i.e.,
$\mathbf{E}  \geq \mathbf{0}$ and $ \mathbf{A}  \geq
\mathbf{0},~\mathbf{1}^{\top}\mathbf{A}=\mathbf{1}$.
\section{Proposed Method}
In our work, we propose an NMF based unmixing framework which uses a
handcrafted regularizer and learnt priors to enhance the unmixing
performance. More specially, we use an $\ell_{2,1}$-norm to enhance the
sparsity of abundances, and we plug the learnt spectral and spatial
information by using a denoising operator. We shall elaborate our framework
as follows.
\subsection{Objective Function}
We use the NMF model to solve the blind unmixing problem. The general
objective function is expressed as:
\begin{equation}\label{eq.loss0}
\mathcal{L}(\mathbf{E}, \mathbf{A})= \mathcal{L}_{\rm data}(\mathbf{E}, \mathbf{A})
+ \alpha \mathcal{L}_{\rm hand}(\mathbf{E},\mathbf{A}) + \mu \mathcal{L}_{\rm learnt}(\mathbf{E}, \mathbf{A}).
\end{equation}
The terms on the right-hand-side of~\eqref{eq.loss0} are interpreted as follows:
\begin{itemize}
\item   $\mathcal{L}_{\rm data}(\mathbf{E}, \mathbf{A})$ is the term
    associated with the data fitting quality. In this work, we set
    $\mathcal{L}_{\rm data}(\mathbf{E}, \mathbf{A})=
    \frac{1}{2}\|\mathbf{R}-\mathbf{E} \mathbf{A}\|_{\text{F}}^{2}$, with
    $\|\cdot\|_{\text{F}}$ being the Frobenius norm.
\item $\mathcal{L}_{\rm learnt}(\mathbf{E}, \mathbf{A})$ is a
    regularization term that \cred{represents} priors learnt from data,
    which does not has an explicit form. Here we only concentrate on
    priors of $\mathbf{A}$ by setting $\mathcal{L}_{\rm
    learnt}(\mathbf{E}, \mathbf{A}) = \Phi(\mathbf{A})$ for illustration
    purpose.
\item $\mathcal{L}_{\rm hand}(\mathbf{E}, \mathbf{A})$ is a handcrafted
    regularization term. Though the learnt regularizer  $\mathcal{L}_{\rm
    learnt}(\mathbf{E}, \mathbf{A})$ is powerful to represent
    spatial-spectral information of the image,   $\mathcal{L}_{\rm
    hand}(\mathbf{E}, \mathbf{A})$ is still necessary to encode physically
    meaningful properties \cred{and the prior knowledge of endmebers} that
    might not easy to be learnt. Here, we use $\mathcal{L}_{\rm
    hand}(\mathbf{E},\mathbf{A})=\|\mathbf{A}\|_{2,1}$, which is a
    structured sparsity regularizer  showing effectiveness in sparse
    unmixing. \cred{Other meaningful handcrafted regularization can also
    be adopted.}
\item $\alpha$ is a positive parameter that controls the impact of the
    sparse regularizer. The positive regularization parameter $\mu$
    controls the strength of plugged priors.
\end{itemize}

Then, the unmixing problem writes:
\begin{equation}\label{eq.loss_2}
\begin{split}
\widehat{\mathbf{E}}, \widehat{\mathbf{A}}=&\arg\min _{\mathbf{E}, \mathbf{A}}
 \frac{1}{2}\|\mathbf{R}-\mathbf{E} \mathbf{A}\|_{\text{F}}^{2}+\alpha\|\mathbf{A}\|_{2,1}+\mu\Phi(\mathbf{{A}}),\\
&\text{s.t.}~~\mathbf{E}  \geq \mathbf{0},~\mathbf{A} \geq \mathbf{0},~\mathbf{1}^{\top}\mathbf{A}=\mathbf{1}, \\
\end{split}
\end{equation}
where $\|\mathbf{A}\|_{2,1}=\sum_{i=1}^{P}\|\mathbf{A}_{i}\|_{2}$, and
$\mathbf{A}_i$ is the $i$-th row of $\mathbf{A}$.  The same as some previous
works~\cite{he2017total,feng2018hyperspectral}, we introduce an auxiliary
variable $\mathbf{\tilde{A}}$ and constraint
$\mathbf{\tilde{A}}=\mathbf{A}$. The objective function~\eqref{eq.loss_2} is
rewritten as:
\begin{equation}\label{eq.loss_3}
\begin{split}
\mathcal{L}(\mathbf{E}, \mathbf{A},\mathbf{\tilde{A}})=
&\min _{\mathbf{E}, \mathbf{A}, \mathbf{\tilde{A}}} \frac{1}{2}\|\mathbf{R}-\mathbf{E} \mathbf{A}\|_{\text{F}}^{2}
+\alpha\|\mathbf{A}\|_{2,1}+\mu\Phi(\mathbf{\tilde{A}}),\\
\text{s.t.}~~\mathbf{E} & \geq \mathbf{0},~\mathbf{A} \geq \mathbf{0},~\mathbf{1}^{\top}\mathbf{A}=\mathbf{1},~\mathbf{\tilde{A}}=\mathbf{A}. \\
\end{split}
\end{equation}
The associated augmented Lagrangian function is given by
\begin{equation}\label{eq.loss_4}
\begin{split}
\mathcal{L}(\mathbf{E}, \mathbf{A}, \mathbf{\tilde{A}})=&\min _{\mathbf{E}, \mathbf{A}, \mathbf{\tilde{A}}}
 \frac{1}{2}\|\mathbf{R}-\mathbf{E} \mathbf{A}\|_{\text{F}}^{2}+\frac{\lambda}{2}\|\mathbf{A}-\mathbf{\tilde{A}}\|_{\text{F}}^{2}\\
&+\alpha\|\mathbf{A}\|_{2,1}+\mu\Phi(\mathbf{\tilde{A}})\\
\text{s.t.}~~\mathbf{E}  &\geq \mathbf{0},~\mathbf{A} \geq \mathbf{0},~\mathbf{1}^{\top}\mathbf{A}=\mathbf{1}, \\
\end{split}
\end{equation}
where $\lambda$ is the penalty parameter. The blind unmixing problem can be
solved by iteratively addressing the following three subproblems:
\begin{align}
\mathbf{E}=&\arg \min _{\mathbf{E}} \mathcal{L}(\mathbf{E}, \mathbf{A}, \mathbf{\tilde{A}})\\
\label{eq.loss_6}\mathbf{A}=&\arg \min _{\mathbf{A}} \mathcal{L}(\mathbf{E}, \mathbf{A}, \mathbf{\tilde{A}})\\
\mathbf{\tilde{A}}=&\arg \min _{\mathbf{\tilde{A}}} \mathcal{L}(\mathbf{E}, \mathbf{A}, \mathbf{\tilde{A}}).
\end{align}
The first two subproblems are forward models used to update the endmember
and abundance matrices. We use the NMF method to solve the first two
subproblems. As to be seen later, the third subproblem can be considered as
an image denoising problem which can be solved by a variety of denoisers.
This framework is flexible and can automatically plug spatial and spectral
priors with the choice of different denoisers.
\subsection{Optimization}
\subsubsection{Endmember estimation}
In order to estimate the endmember matrix, we devote to solve the following
minimization objective problem:
\begin{equation}\label{eq.end}
\mathcal{L}(\mathbf{E})=\min_{\mathbf{E}} \frac{1}{2}\|\mathbf{R}-\mathbf{E} \mathbf{A}\|_{\text{F}}^{2}+\text{Tr}(\mathbf{\Lambda} \mathbf{E}),
\end{equation}
where $\mathbf{\Lambda}$ is the Lagrange multiplier to control the impact of
ENC. We calculate the gradients about $\mathbf{E}$ and set it to
$\mathbf{0}$:
\begin{equation}\label{eq.end_1}
  \frac{\partial \mathcal{L}(\mathbf{E})}{\partial\mathbf{E}}=\mathbf{EA}\mathbf{A}^{\top}-\mathbf{R}\mathbf{A}^{\top}+\mathbf{\Lambda}=\mathbf{0}.
\end{equation}
By element-wise multiplication $\mathbf{E}$ of both sides
of~\eqref{eq.end_1} and according to the Karush-Kuhn-Tucker (K-K-T)
conditions $\mathbf{E}\odot\mathbf{\Lambda}=\mathbf{0}$, the endmember
matrix $\mathbf{E}$ is updated as:
\begin{equation}\label{eq.end_2}
  \mathbf{E}\leftarrow \mathbf{E} \odot(\mathbf{RA}^{\top})\oslash(\mathbf{EAA}^{\top}),
\end{equation}
where $\odot$ is element-wise multiplication, and $\oslash$ is element-wise division.
\subsubsection{Abundance estimation}
The second subproblem is to estimate abundance. It is difficult to solve
ASC, and we use two augmented matrixes $\mathbf{R}_{f}, \mathbf{E}_{f}$ to
address this issue:
\begin{equation}\label{eq.abu}
\mathbf{R}_{f}= \left[
   \begin{array}{c}
     \mathbf{R} \\
     \delta\mathbf{1}_{N}^{\top} \\
   \end{array}
 \right],
~~ \mathbf{E}_{f}= \left[
   \begin{array}{c}
     \mathbf{E} \\
     \delta\mathbf{1}_{P}^{\top} \\
   \end{array}
 \right],
\end{equation}
where $\delta$ is the penalty parameter controlling the strength of ASC. The
objective problem of abundance estimation is as follows:
\begin{equation}\label{eq.abu_1}
\begin{split}
\mathcal{L}(\mathbf{A})=&  \min_{\mathbf{A}} \frac{1}{2}\|\mathbf{R}_{f}-\mathbf{E}_{f}
 \mathbf{A}\|_{\text{F}}^{2}+\frac{\lambda}{2}\|\mathbf{A}-\mathbf{\tilde{A}}\|_{\text{F}}^{2}+\\
&\text{Tr}(\mathbf{\Gamma} \mathbf{A})+\alpha\|\mathbf{A}\|_{2,1},
\end{split}
\end{equation}
where $\mathbf{\Gamma}$ is the Lagrange multiplier to control the impact of ANC.
We calculate the derivative of~\eqref{eq.abu_1} and set it to 0:
\begin{equation}\label{eq.abu_2}
  \frac{\partial \mathcal{L}(\mathbf{A})}{\partial\mathbf{A}}=\mathbf{E}_{f}^{\top}\mathbf{E}_{f}\mathbf{A}-\mathbf{E}_{f}\mathbf{R}_{f}
  +\lambda(\mathbf{A}-\mathbf{\tilde{A}})+\mathbf{\Gamma}+\alpha \mathbf{DA}=\mathbf{0},
\end{equation}
where
\begin{equation}
\mathbf{D}=\operatorname{diag}\left\{\frac{1}{\left\|\mathbf{A}_{1}\right\|_{2}},
 \frac{1}{\left\|\mathbf{A}_{2}\right\|_{2}}, \cdots, \frac{1}{\left\|\mathbf{A}_{P}\right\|_{2}}\right\}.
\end{equation}
According to KKT conditions, we get
$\mathbf{A}\odot\mathbf{\Gamma}=\mathbf{0}$. By element-wise multiplication
$\mathbf{A}$ of both sides of~\eqref{eq.abu_2}, we update the abundance
matrix as follows:
\begin{equation}\label{eq.abu_3}
  \mathbf{A}\leftarrow \mathbf{A}\odot (\mathbf{E}_{f}\mathbf{R}_{f}
  +\lambda\mathbf{\tilde{A}})\oslash[\mathbf{E}_{f}^{\top}\mathbf{E}_{f}\mathbf{A}+\lambda \mathbf{A}+\alpha \mathbf{DA}].
\end{equation}
\subsubsection{Plugged priors}
In the third subproblem, we focus on solving the following optimization problem:
\begin{equation}\label{eq.denoise}
\mathcal{L}(\mathbf{\tilde{A}})= \frac{\lambda}{2}\|\mathbf{A}-\mathbf{\tilde{A}}\|_{\text{F}}^{2}+\mu\Phi(\mathbf{\tilde{A}}).
\end{equation}
This step can be seen as an abundance denoising problem, where
$\mathbf{\tilde{A}}$ is the clean version of abundance maps. This problem
can be rewritten as:
\begin{equation}\label{eq.denoise_1}
\mathcal{L}(\mathbf{\tilde{A}})= \frac{1}{2(\sqrt{\mu/\lambda})^2}
\|\mathbf{A}-\mathbf{\tilde{A}}\|_{\text{F}}^{2}+\Phi(\mathbf{\tilde{A}}).
\end{equation}
According to the maximum a posteriori model (MAP), the problem
in~\eqref{eq.denoise_1} can be regarded as denoising abundance maps with
additive gaussian noise with a standard deviation
$\sigma_{n}=\sqrt{\mu/\lambda}$ and the priors are encoded in
$\Phi(\mathbf{\tilde{A}})$. In our proposed unmixing framework, instead of
solving this problem using optimization methods, we use denoisers to solve
this regularized optimization problem. Plugged denoisers can automatically
carry prior information. The denoising operator is actually performed in the
3D domain to fully exploit the spectral and spatial information, we
rewrite~\eqref{eq.denoise_1} as follows:
\begin{equation}\label{eq:denosier_2}
\mathbf{\tilde{A}} \leftarrow Denoiser(\mathcal{T}(\mathbf{A}), \sigma_{n}),
\end{equation}
where $\mathcal{T}(\cdot)$ is an operator that transforms the 2D matrix to
3D data cube. Plenty of denoisers can be applied in this step and carry
various kinds of priors. In our work, we use a conventional linear denoiser
non-local means denoising (\texttt{NLM}), a nonlinear denoiser
block-matching and 3D filtering (\texttt{BM3D}) to our NMF based unmixing
framework. These two denoisers are 2D cube based and solve the abundances
denoising problem band by band. Further, two 3D cube based denoisers
\texttt{BM4D} and a total variation regularized low-rank tensor
decomposition denoising model (\texttt{LRTDTV}) are also plugged into our
proposed framework. We denote our proposed methods as PNMF-NLM, PNMF-BM3D,
PNMF-BM4D and PNMF-LRTDTV, respectively. Our NMF based unmixing framework is
presented in Algorithm~\ref{alg_1}.

\begin{algorithm}[!t]
	\label{alg_1}
	\KwIn{Hyperspectral image $\bf R$, regularization para -meters $\lambda$, $\alpha$ $\mu$, $\delta$, the iteration number $K$.}
	\KwOut{Endmember matrix $\bf E$, abundance matrix $\bf A$.}
	Initialize ${\bf E}$ with VCA, $\bf A$, $\bf \tilde{A}$ with FCLS. \\
	\While{Stopping criterias are not met and $k\le K$}{
			Update $\mathbf{E}_k$ with~\eqref{eq.end_2};\\
            Augment $\bf R$ and $\bf E$ to obtain $\mathbf{R}_f$ and $\mathbf{E}_f$, respectively;\\
            Update $\mathbf{A}_k$ with~\eqref{eq.abu_3};\\
            Update $\mathbf{\tilde{A}}_k$ using the strategy of~\eqref{eq:denosier_2};\\
            $k=k+1$;
	}
	\caption{Proposed NMF based framework for hyperspectral unmixing.}
\end{algorithm}

\section{Experiments}

\begin{table*}[]
\footnotesize \centering
\caption{\small RMSE, SAD and PSNR Comparison of Synthetic Dataset.}
\vspace{-0.2cm}
\renewcommand\arraystretch{1.5}
\begin{tabular}{c|c|c|c|c|c|c|c|c|c|c|c|c}
\hline
\hline
                       & \multicolumn{3}{c|}{5dB}                             & \multicolumn{3}{c|}{10dB}                            & \multicolumn{3}{c|}{20dB}                            & \multicolumn{3}{c}{30dB}                            \\ \hline
\multicolumn{1}{l|}{} & RMSE            & SAD             & PSNR             & RMSE            & SAD             & PSNR             & RMSE            & SAD             & PSNR             & RMSE            & SAD             & PSNR             \\ \hline
VCA-SUnSAL-TV          & 0.1392          & 6.0675          & 17.5317          & 0.0692          & 3.0419          & 23.1989          & 0.0240          & 0.7611          & 32.4014          & 0.0094          & 0.1880          & 40.5592          \\ \hline
CoNMF                  & 0.1074          & 6.8326          & 9.7881           & 0.0773          & 4.4600          & 22.2385          & 0.0248          & 0.7609          & 32.1133          & 0.0094          & 0.1879          & 40.5005          \\ \hline
TV-RSNMF                & 0.1390          & 6.0034          & 9.6711           & 0.0740          & 4.3929          & 22.6155          & 0.0265          & 1.0429          & 31.5214          & 0.0092          & 0.2169          & 40.7600          \\ \hline
NMF-QMV                & 0.1392          & 23.4638         & 17.1248          & 0.0854          & 8.3570          & 21.3688          & 0.0247          & 0.6208          & 32.1312          & 0.0092          & 0.1902          & 40.6972          \\ \hline
PNMF-NLM               & 0.0817          & 5.1458          & 22.2549          & 0.0560          & 2.9961          & 24.8780          & 0.0212          & 0.5116          & 33.0728          & 0.0092          & 0.1877          & 40.1695          \\ \hline
PNMF-BM3D              & 0.0779          & 5.1164      & \textbf{23.4658}          & \textbf{0.0557} & \textbf{2.0428} & 25.2735          & \textbf{0.0176} & \textbf{0.3898} & \textbf{35.2867}          & 0.0092          & 0.1878          & 40.5740          \\ \hline
PNMF-BM4D              & 0.0838          & 5.2499          & 21.2835          & 0.0579          & 2.0840          & \textbf{25.4650} & 0.0186          & 0.4242          & 34.4498          & 0.0092          & \textbf{0.1876} & \textbf{41.0646} \\ \hline
PNMF-LRTDTV            & \textbf{0.0773} & \textbf{5.0415} & {21.9143} & 0.0597          & 2.7060          & 25.1639          & 0.0211          & 0.4873          & {33.5669} & \textbf{0.0091} & 0.1878          & 40.5740          \\ \hline\hline
\end{tabular}
\label{tab.syn_results}
\vspace{-3mm}
\end{table*}


In this section, we use both synthetic data and real data experiments to
evaluate the unmixing performance of our proposed NMF based unmixing
framework. Our methods were compared with several state-of-the-art methods.
First, we considered a sequential unmixing method, where we extracted the
endmembers with VCA~\cite{Nascimento2003Vertex} and estimated the abundances
using SUnSAL-TV~\cite{iordache2012total}. We also compared with NMF based
methods. CoNMF~\cite{li2016robust} is a robust collaborative NMF method for
hyperspectral unmixing. TV-RSNMF~\cite{he2017total} is a TV regularized
reweighted sparse NMF method. NMF-QMV~\cite{Lina2019Regularization} is a
variational minimum volume regularized NMF method. We used the spectral
angle distance (SAD) to evaluate the endmember extraction results: $
  \text{SAD} = \frac{1}{P}\sum_{k=1}^{P}\cos^{-1}\left(\frac{\mathbf{e}_{k}^{\top}\hat{\mathbf{e}}_{k}}{\|\mathbf{e}_{k}\|\|\hat{\mathbf{e}}_{k}\|}\right),
$ where $\mathbf{e}_{k}$ is the ground-truth and $\hat{\mathbf{e}}_{k}$ is
the estimated endmember. We used the root mean square error (RMSE) to
evaluate the abundance estimation results: $
  \text{RMSE} = \sqrt{\frac{1}{NP}\sum_{i=1}^{N}\|\mathbf{a}_{i}-\hat{\mathbf{a}}_{i}\|^{2}},
$ where $\mathbf{a}_{i}$ is the $i$-th column of $\mathbf{A}$ and represents
the ground-truth, and $\hat{\mathbf{a}}_{i}$ is the estimated abundance.
Further, we used the peak signal-to-noise ratio (PSNR) to evaluate the
denoising quality between estimated abundances and ground-truth: $
  \text{PSNR} = 10\times \log_{10}\left(\frac{\text{MAX}^{2}}{\text{MSE}}\right),
$ in which $\text{MAX}$ is the maximum abundance value, and
$\text{MSE}=\frac{1}{N}\sum_{i}\sum_{j}[A(i,j)-\hat{A}(i,j)]^{2}$, where
$\hat{A}$ is the estimated abundance and $A$ is the clean ground-truth.
\subsection{Synthetic data}
In this experiment, we generated the synthetic data using Hyperspectral
Imagery Synthesis tools with Gaussian
Fields\footnote{\cred{http://www.ehu.es/ccwintco/index.php/Hyperspectral\_Imagery\_Synthesis\_
tools\_for\_MATLAB}}, \cred{and the LMM is adopted}. The spatial size of the
synthetic data is $256\times 256$.
A selection of four endmembers from USGS spectral library were used as the
endmember library, \cred{with $224$ bands covering wavelength from 400~nm to
2500~nm}. \cred{There are both mixed and pure pixels in the dataset.} To
evaluate the robustness of our methods and the efficacy of plugged
denoisers, we added Gaussian noise to the clean data with the
signal-to-noise ratio (SNR) setting to 5~dB, 10~dB, 20~dB and 30~dB.

\begin{figure*}
  \centering
  \includegraphics[width=17cm]{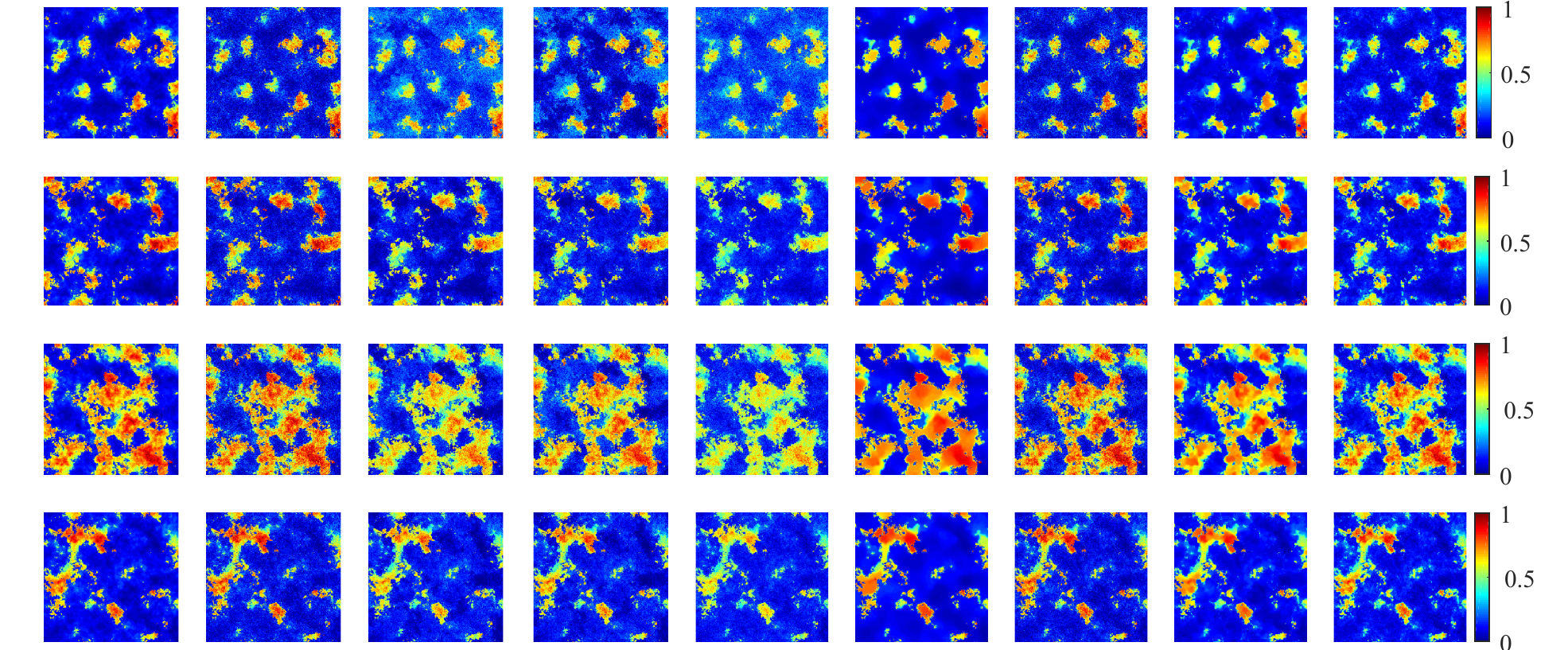}\\
  \caption{\cred{Abundance maps of synthetic data (SNR = 10~dB). From top to bottom:
  different endmembers. From left to right: Ground-truth, VCA-SUnSAL-TV, CoNMF, TV-RSNMF, NMF-QMV,
PNMF-NLM, PNMF-BM3D, PNMF-BM4D and PNMF-LRTDTV.}}\label{fig.map_syn}
\end{figure*}

\begin{figure}
  \centering
  \includegraphics[width=9cm]{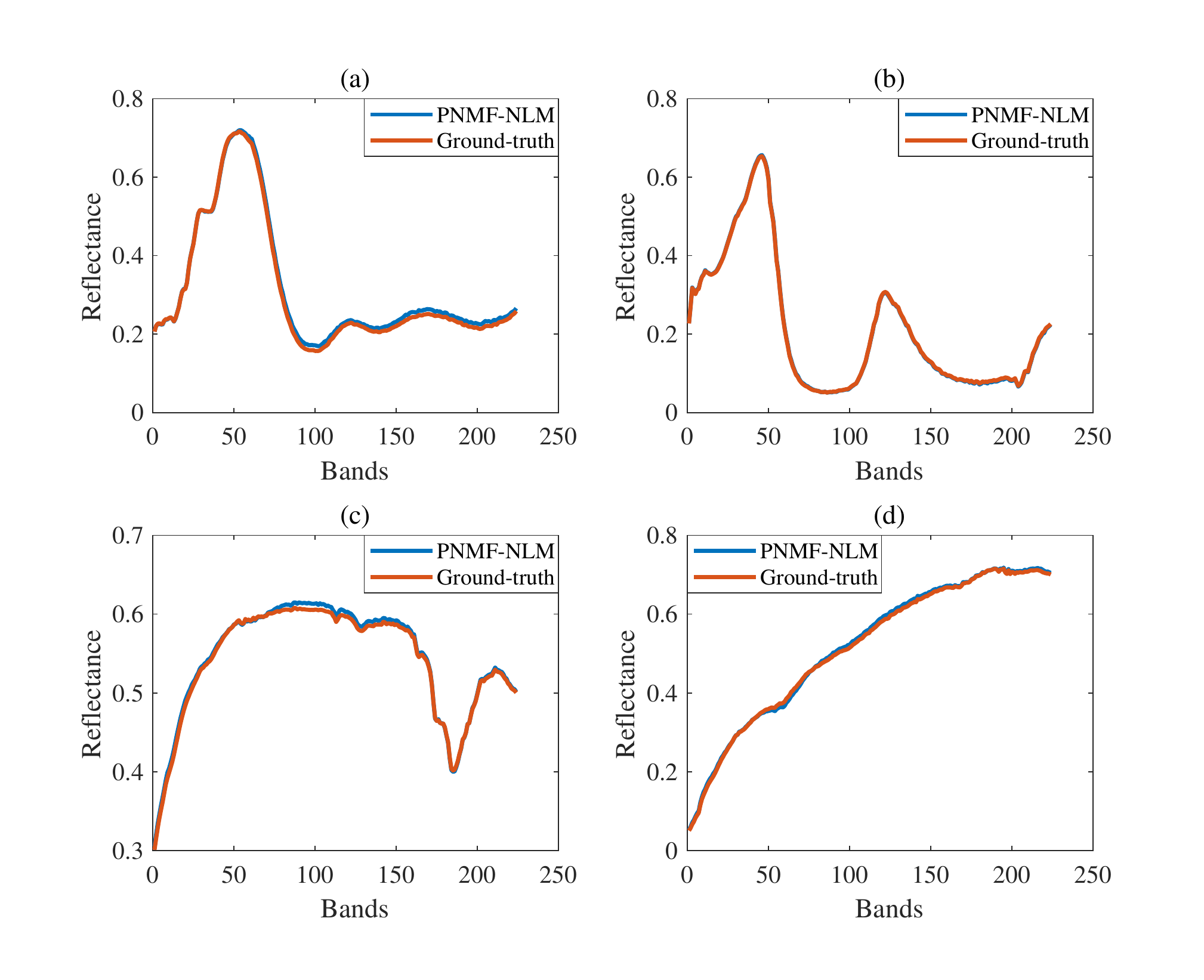}\\
  \caption{\cred{Endmembers extracted by PNMF-NLM (SNR=20~dB).}}\label{fig.end_syn}
\end{figure}

\begin{figure}[h]
  \centering
  \includegraphics[width=8cm]{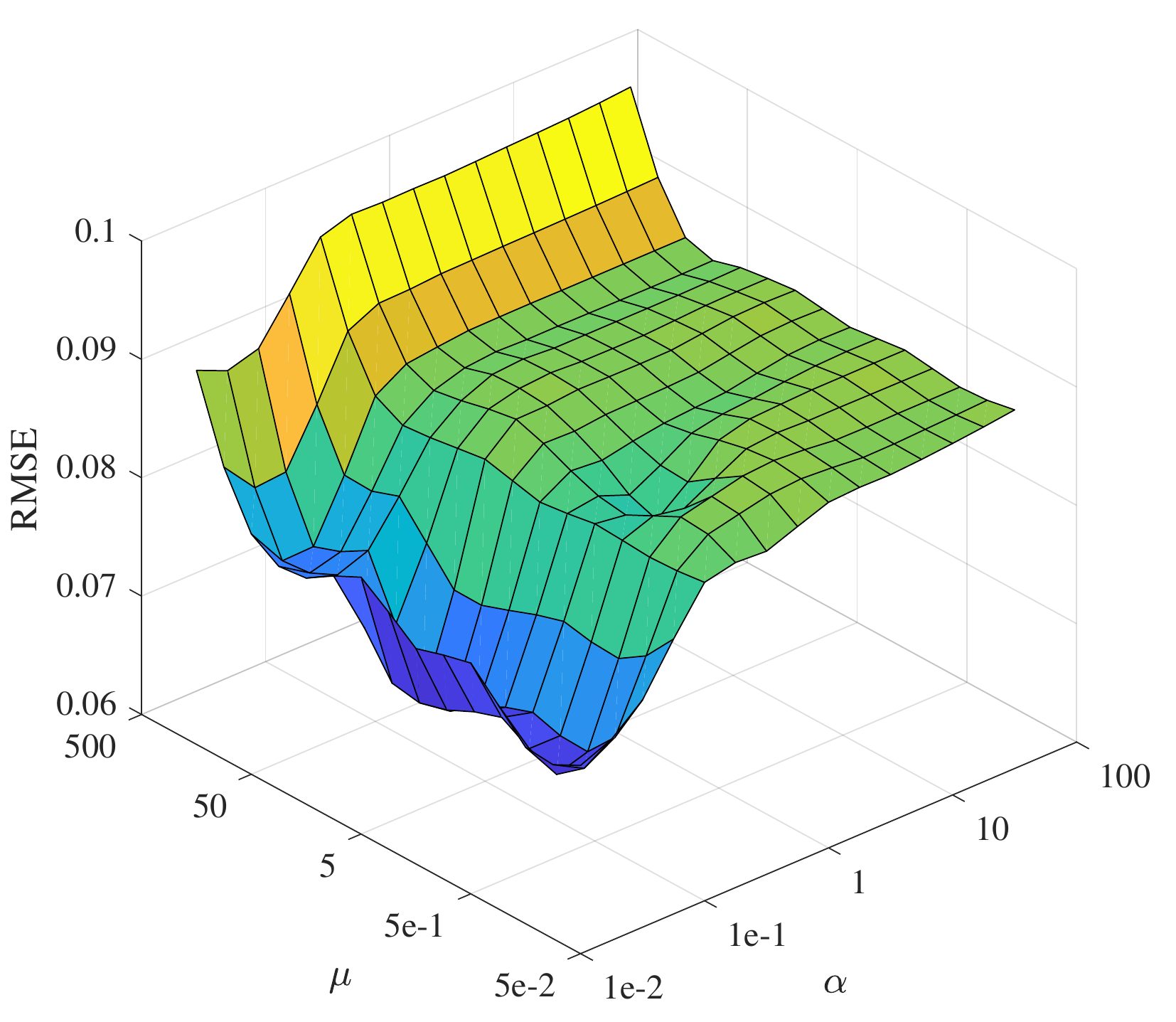}\\
  \caption{\cred{RMSE as a function of the regularization parameters for PNMF-NLM.}}\label{fig.RMSE_curve}
\end{figure}

\begin{figure}[h]
  \centering
  \includegraphics[width=7.8cm]{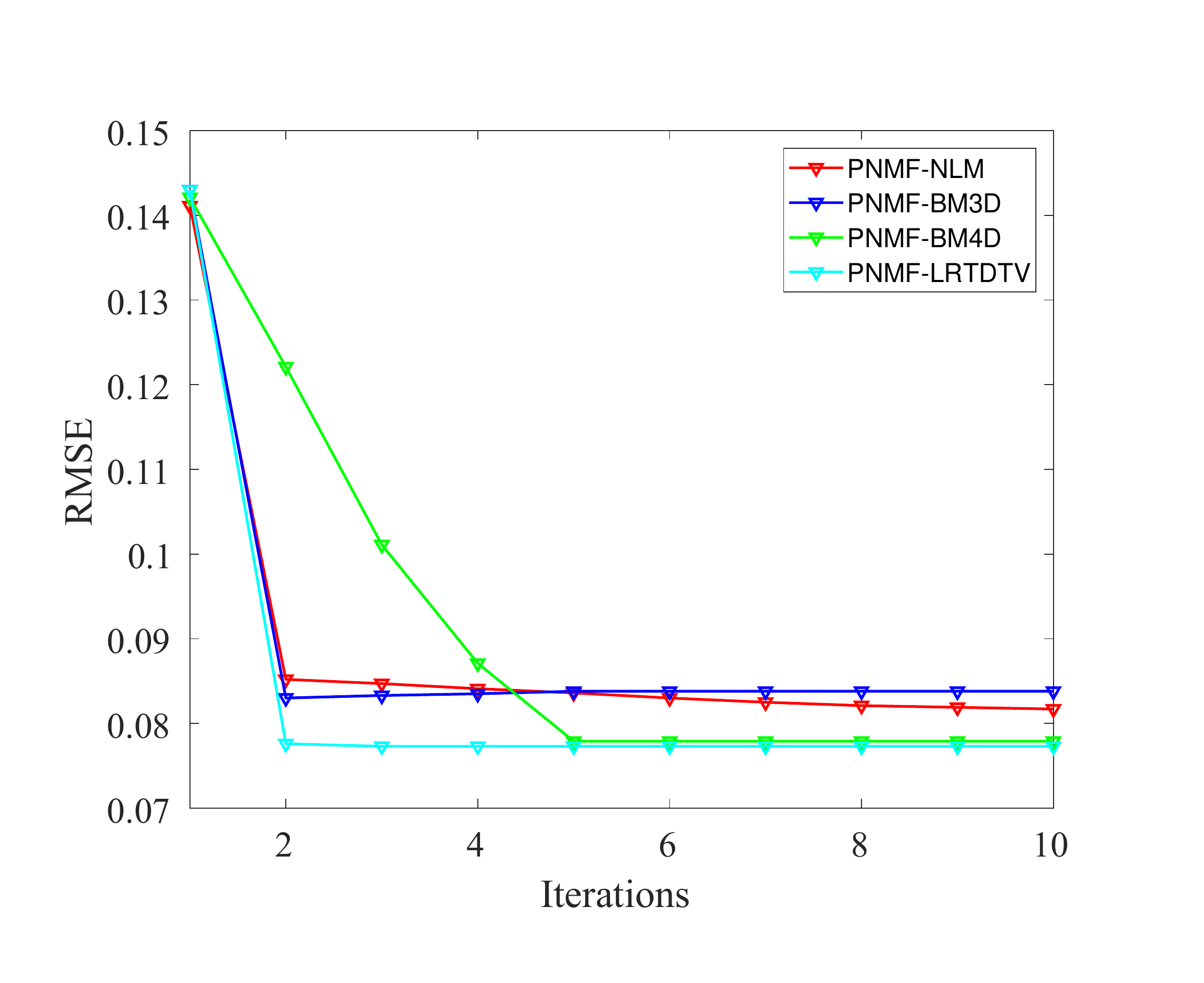}\\
  \caption{The RMSE convergence curves of synthetic data of our proposed methods (5dB).}\label{fig.loss_curve}
\end{figure}

In our work, we used VCA to initialize the endmembers and the fully
constrained least square method (FCLS)~\cite{heinz2001fully} to initialize
the abundances. After multiple experiments, we set the penalty parameter of
$\ell_{2,1}$ ($\alpha$) to 0.1, the penalty parameter of denoising term
($\lambda$) to $3\times 10^{4}$, the penalty parameter of ASC ($\rho$)
\cred{to} 10, and $\mu$ to 500. Unmixing results of RMSE, SAD and PSNR of
this experiment are reported in Table~\ref{tab.syn_results}. From the
results, we observe that our proposed methods get \cred{the best RMSE and
PSNR results} and achieve the lowest mean SAD values than other methods.
This highlights the effects of our sparse regularizer and the superiority of
priors learnt by denoisers. \cred{Moreover, benefitting from the denoisers,
when the noise level is high, the unmixing results of our methods show more significant
 enhancement. This indicates that our methods are robust to noise.}
\cred{Figure~\ref{fig.map_syn} shows the abundance maps of four compared
methods and our proposed methods with SNR = 10~dB. We can see that the
abundance estimated results of our methods are with less noise and more
close to ground-truth.} \cred{The endmember extracted results of PNMF-NLM
(SNR = 20~dB) are shown in Figure~\ref{fig.end_syn}, where the red curves
are ground-truth, and the blue curves are estimated endmembers.}
\cred{Moreover, Figure~\ref{fig.RMSE_curve} presents how parameters affect
the unmixing results with SNR = 10~dB.} The RMSE convergence curves of
synthetic data of our proposed methods are shown in
Figure~\ref{fig.loss_curve}, which indicate that our NMF based unmixing
framework has a stable convergence property.

\subsection{Real data}
\subsubsection{Cuprite dataset}
In this experiment, we used a well-known real hyperspectral dataset (AVIRIS
Cuprite) to evaluate our unmixing methods. The dataset was captured from the
Cuprite mining district in west-central Nevada by AVIRIS in 1997. We used a
subimage of size $250\times 191$ in our experiment. This dataset has $224$
bands. Following other works~\cite{li2016robust,tong2020adaptive}, we
removed the water absorption and noisy bands (2, 105-115, 150-170, 223 and
224) with 188 exploitable bands remained. The number of endmembers was set
to 12.

\begin{figure}
  \centering
  \includegraphics[width=7.5cm]{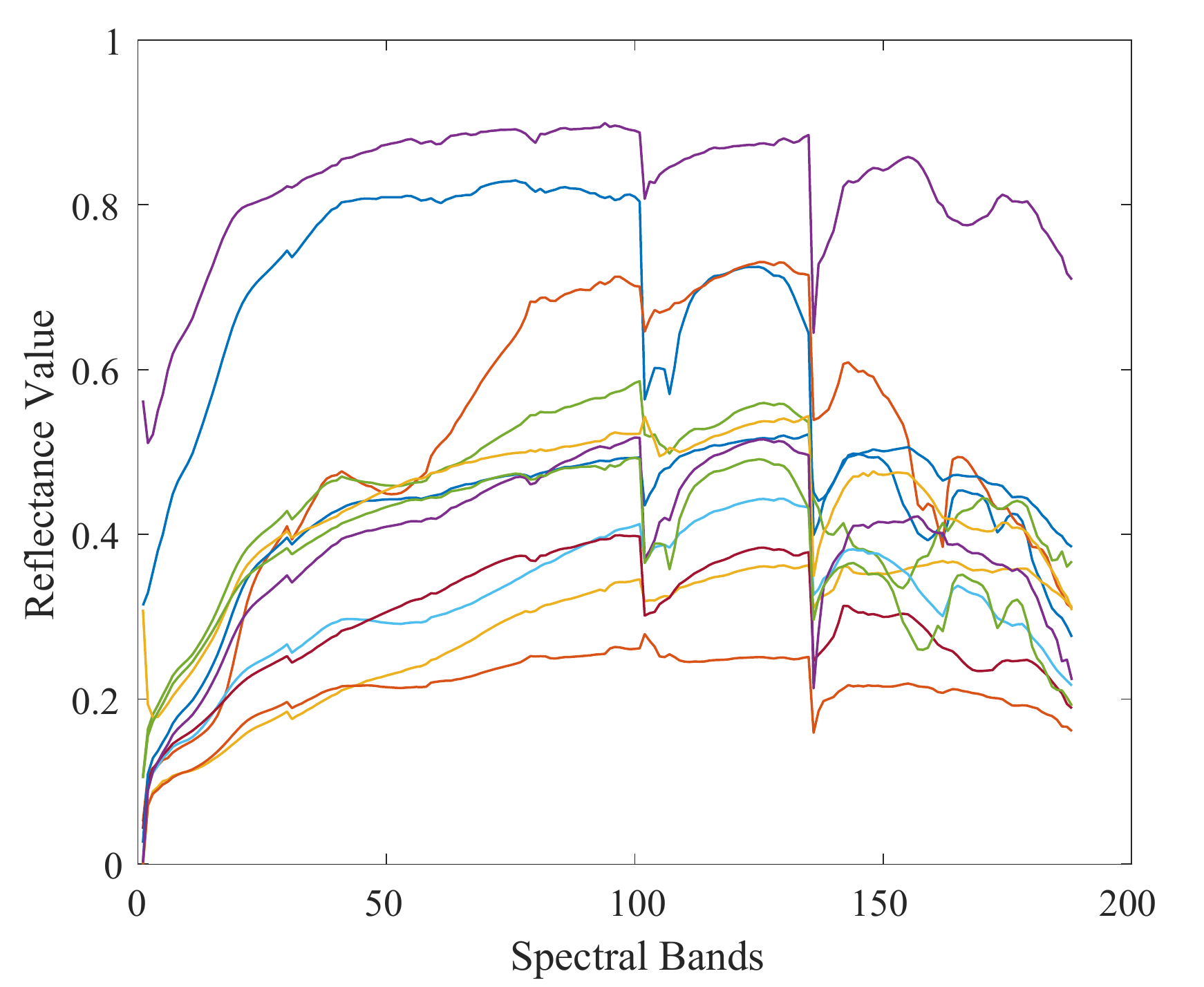}\\
   \vspace{-0.3cm}
  \caption{Twelve endmembers extracted from PNMF-BM4D of Cuprite dataset.}\label{fig.spe_curve}
    \vspace{-0.3cm}
\end{figure}

\begin{figure*}[t]
  \centering
  \includegraphics[width=16.5cm]{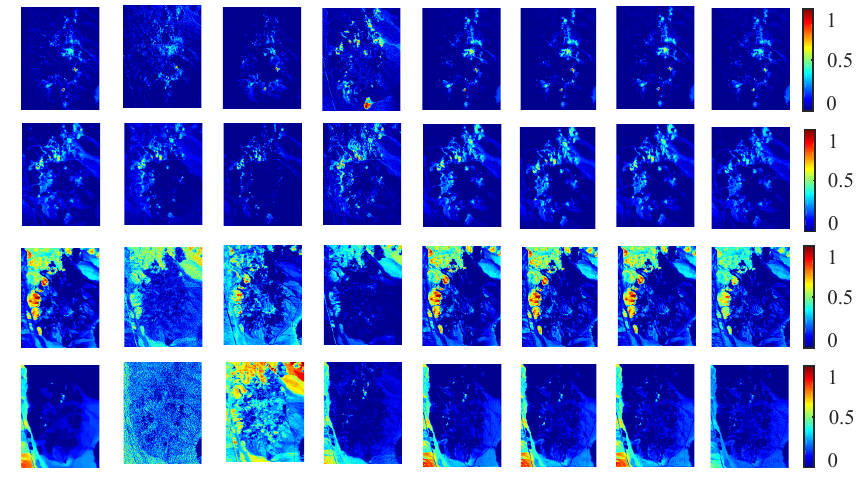}\\
  \vspace{-0.4cm}
  \caption{\cred{Abundance maps of Cuprite data. From top to bottom: selected endmembers. From left to right:
  VCA-SUnSAL-TV, CoNMF, TV-RSNMF,  NMF-QMV,
   PNMF-NLM, PNMF-BM3D, PNMF-BM4D and PNMF-LRTDTV.}}\label{fig.real_abundance}
   \vspace{-0.3cm}
\end{figure*}

In this experiment, we set $\alpha$ to 0.1, $\lambda$ to $3\times 10^{4}$,
$\rho$ to 10 \cred{and} $\mu$ to 100. As there is no ground-truth for this
dataset, we can not give a quantitative comparison. Like many previous
works, we evaluate the unmixing results in an intuitive manner. The
abundance maps of selected \cred{materials} of the Cuprite data are shown in
Figure~\ref{fig.real_abundance}. \cred{We observe that our proposed methods
provide clearer and sharper results with several locations emphasized and
more detailed information.
The endmember extraction results of PNMF-BM4D are shown in Figure~\ref{fig.spe_curve}.
We also used reconstructed error (RE) to qualify these methods:}
$
  \text{RE}=\sqrt{\frac{1}{NP}\sum_{i=1}^{N}\parallel \mathbf{r}_{i}-\mathbf{\hat{r}}_{i}\parallel^{2}},
$ where $\hat{\mathbf{r}}_{i}$ represents the reconstructed pixel, and
$\mathbf{r}_{i}$ is the ground-truth.
The RE comparison is shown in Table~\ref{tab.RE_results}.
We observe that PNMF-BM4D gets
the lowest RE results. Note that without the ground-truth information
on the abundances and endmembers for the real data, RE is not necessarily proportional
to the quality of unmixing performance, and therefore it can only be
considered as complementary information. Note that our methods use denoisers
to plug learnt priors, which can also denoise the abundance maps but may
increase the RE compared to the noisy input.

\begin{table*}
\footnotesize \centering
\caption{\small RE Comparison of Cuprite Dataset.}
\vspace{-0.2cm}
\renewcommand\arraystretch{1.2}
\begin{tabular}{ccccccccc}\hline\hline
Algorithm & VCA-SUnSAL-TV & CoNMF  & TV-RSNMF & NMF-QMV & PNMF-NLM & PNMF-BM3D & PNMF-BM4D       & PNML-LRTDTV \\
RE        & 0.0087        & 0.0235 & 0.0085   & 0.0425  & 0.0083   & 0.0081    & \textbf{0.0070} & 0.0091\\\hline\hline
\end{tabular}
\label{tab.RE_results}
\end{table*}
\subsubsection{Jasper Ridge dataset}
The Jasper Ridge dataset with $100\times100$ pixels was used for this
purpose. The data consist of 224 spectral bands ranging from 380~nm to
2500~nm with spectral resolution up to 10~nm. After removing channels [1-3,
108-112, 154-166 and 220-224] affected by dense water vapor and the
atmosphere, 198 channels were remained. Four prominent endmembers existing
in this data are considered in our experiments.

\begin{figure*}[t]
  \centering
  \includegraphics[width=17cm]{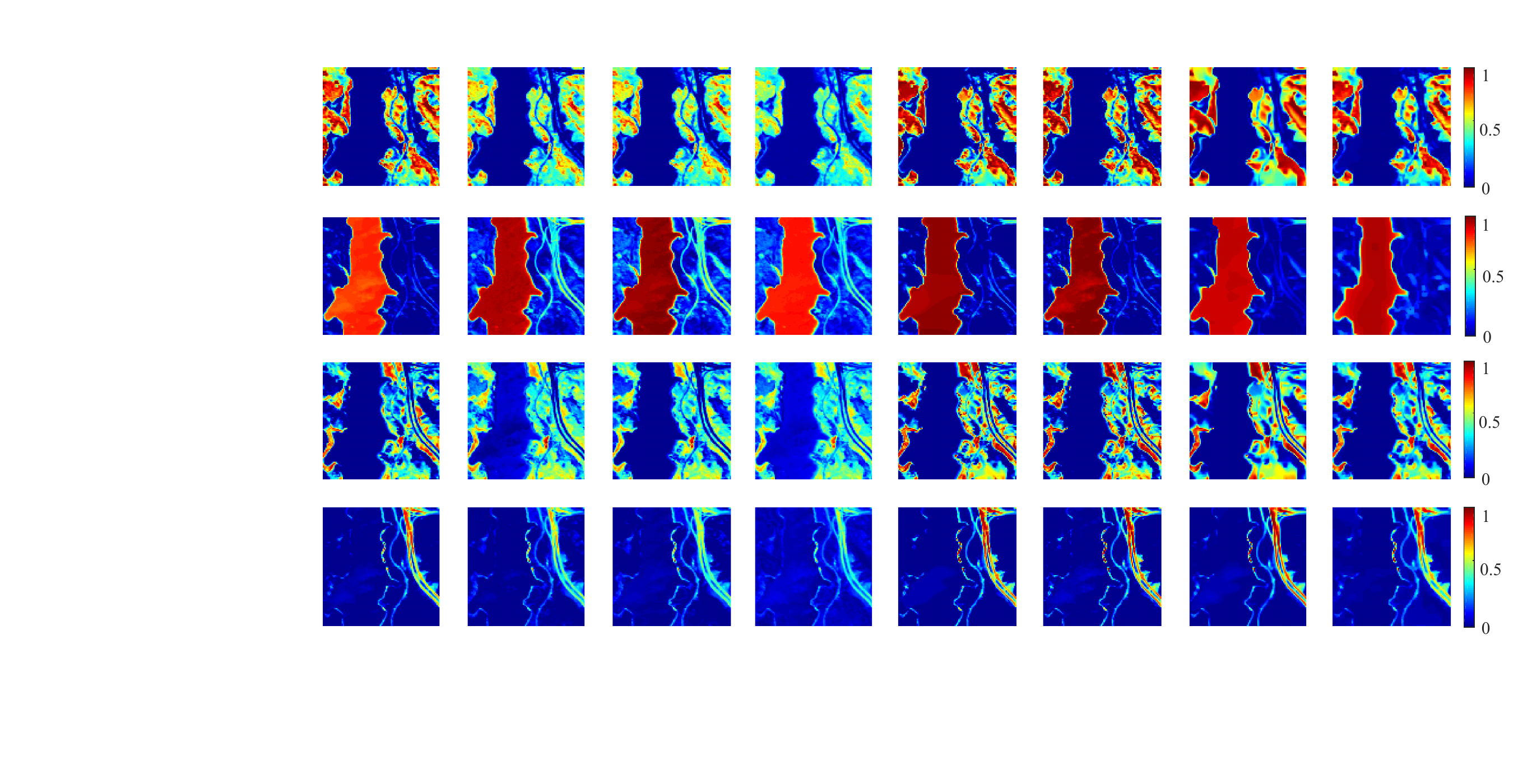}\\
  \caption{Abundance maps of Jasper Ridge data. From top to bottom: four endmembers. From left to right: VCA-SUnSAL-TV, CoNMF, TV-RSNMF, NMF-QMV,
PNMF-NLM, PNMF-BM3D, PNMF-BM4D and PNMF-LRTDTV.}\label{fig.map_jas}
\end{figure*}

\begin{figure}[h]
  \centering
  \includegraphics[width=7.5cm]{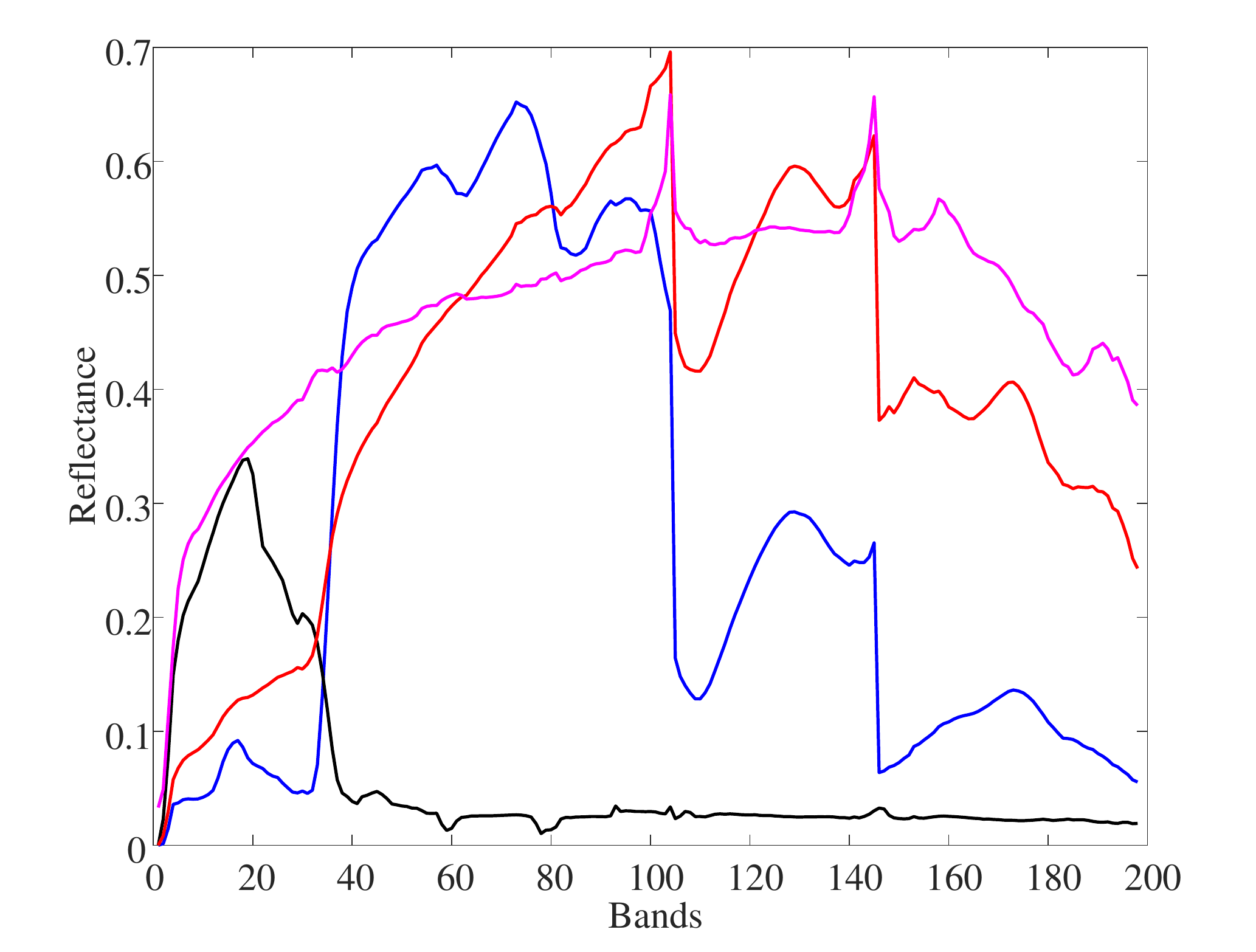}\\
  \caption{Four endmembers extracted from PNMF-BM4D of Jasper Ridge dataset.}\label{fig.curve_jas}
\end{figure}

The experiment settings are
the same as Cuprite dataset. We set $\alpha$ to 0.1, $\lambda$ to
$3\times10^{-4}$, $\rho$ to 10 and $\mu$ to 100. The abundance estimated
results are shown in Figure~\ref{fig.map_jas}, and the endmembers extracted
from PNMF-BM4D is shown in~\ref{fig.curve_jas}. The unmixing results
indicate that our proposed methods are with less noise and more smoothness.
RE results of Jasper Ridge dataset are shown in Table~\ref{tab.RE_results_1},
which also demonstrate the effective performance of our methods.


\begin{table*}
\footnotesize \centering
\caption{\small RE Comparison of Jasper Ridge Dataset.}
\vspace{0.2cm}
\renewcommand\arraystretch{1.2}
\begin{tabular}{ccccccccc}\hline\hline
Algorithm & VCA-SUnSAL-TV & CoNMF  & TV-RSNMF & NMF-QMV & PNMF-NLM & PNMF-BM3D & PNMF-BM4D       & PNML-LRTDTV \\
RE        & 0.0257        & 0.0198 & 0.0122   & 0.0434 & \textbf{0.0111}   & 0.0123    & {0.0135} & 0.0117\\
\hline\hline
\end{tabular}
\label{tab.RE_results_1}
\end{table*}

\section{Conclusion}
In this paper, we proposed an NMF based unmixing framework that jointly
\cred{used} handcrafted and learnt regularizers. \cred{We used a denoiser of
abundances to plug learnt priors. Our framework allowed plugging various
denoisers which is flexible and extendable. For illustrating the integration of handcrafted
regularization, we added a structured sparse regularizer to the objective
function to enhance the sparsity of unmixing results.} Experiment results
showed the effectiveness of our proposed methods. \cred{Future work will
focus on the adaptive  parameter selection.}

\bibliographystyle{IEEEtran}
\bibliography{IEEEfull,BIB_Min}\ 

\begin{thebibliography}{10}
\providecommand{\url}[1]{#1}
\csname url@samestyle\endcsname
\providecommand{\newblock}{\relax}
\providecommand{\bibinfo}[2]{#2}
\providecommand{\BIBentrySTDinterwordspacing}{\spaceskip=0pt\relax}
\providecommand{\BIBentryALTinterwordstretchfactor}{4}
\providecommand{\BIBentryALTinterwordspacing}{\spaceskip=\fontdimen2\font plus
\BIBentryALTinterwordstretchfactor\fontdimen3\font minus
  \fontdimen4\font\relax}
\providecommand{\BIBforeignlanguage}[2]{{%
\expandafter\ifx\csname l@#1\endcsname\relax
\typeout{** WARNING: IEEEtran.bst: No hyphenation pattern has been}%
\typeout{** loaded for the language `#1'. Using the pattern for}%
\typeout{** the default language instead.}%
\else
\language=\csname l@#1\endcsname
\fi
#2}}
\providecommand{\BIBdecl}{\relax}
\BIBdecl

\bibitem{keshava2002spectral}
N.~Keshava and J.~F. Mustard, ``Spectral unmixing,'' \emph{IEEE Signal Process.
  Mag.}, vol.~19, no.~1, pp. 44--57, 2002.

\bibitem{bioucas2012hyperspectral}
J.~M. Bioucas-Dias, A.~Plaza, N.~Dobigeon, M.~Parente, Q.~Du, P.~Gader, and
  J.~Chanussot, ``Hyperspectral unmixing overview: Geometrical, statistical,
  and sparse regression-based approaches,'' \emph{IEEE J. Sel. Top. Appl. Earth
  Observat. Remote Sens.}, vol.~5, no.~2, pp. 354--379, 2012.

\bibitem{li2016robust}
J.~Li, J.~M. Bioucas-Dias, A.~Plaza, and L.~Liu, ``Robust collaborative
  nonnegative matrix factorization for hyperspectral unmixing,'' \emph{IEEE
  Trans. Geosci. Remote Sens.}, vol.~54, no.~10, pp. 6076--6090, 2016.

\bibitem{lu2013double}
X.~Lu, H.~Wu, and Y.~Yuan, ``Double constrained nmf for hyperspectral
  unmixing,'' \emph{IEEE Trans. Geosci. Remote Sens.}, vol.~52, no.~5, pp.
  2746--2758, 2013.

\bibitem{salehani2017smooth}
Y.~E. Salehani and S.~Gazor, ``Smooth and sparse regularization for nmf
  hyperspectral unmixing,'' \emph{IEEE J. Sel. Top. Appl. Earth Observat.
  Remote Sens.}, vol.~10, no.~8, pp. 3677--3692, 2017.

\bibitem{wang2016hypergraph}
W.~Wang, Y.~Qian, and Y.~Tang, ``Hypergraph-regularized sparse nmf for
  hyperspectral unmixing,'' \emph{IEEE J. Sel. Top. Appl. Earth Observat.
  Remote Sens.}, vol.~9, no.~2, pp. 681--694, 2016.

\bibitem{lu2019subspace}
X.~Lu, L.~Dong, and Y.~Yuan, ``Subspace clustering constrained sparse nmf for
  hyperspectral unmixing,'' \emph{IEEE Trans. Geosci. Remote Sens.}, 2019.

\bibitem{he2017total}
W.~He, H.~Zhang, and L.~Zhang, ``Total variation regularized reweighted sparse
  nonnegative matrix factorization for hyperspectral unmixing,'' \emph{IEEE
  Trans. Geosci. Remote Sens.}, vol.~55, no.~7, pp. 3909--3921, 2017.

\bibitem{sigurdsson2016blind}
J.~Sigurdsson, M.~O. Ulfarsson, and J.~R. Sveinsson, ``Blind hyperspectral
  unmixing using total variation and $\ell_q $ sparse regularization,''
  \emph{IEEE Trans. Geosci. Remote Sens.}, vol.~54, no.~11, pp. 6371--6384,
  2016.

\bibitem{feng2018hyperspectral}
X.~Feng, H.~Li, J.~Li, Q.~Du, A.~Plaza, and W.~Emery, ``Hyperspectral unmixing
  using sparsity-constrained deep nonnegative matrix factorization with total
  variation,'' \emph{IEEE Trans. Geosci. Remote Sens.}, vol.~56, no.~10, pp.
  6245--6257, 2018.

\bibitem{feng2019hyperspectral}
X.~Feng, H.~Li, and R.~Wang, ``Hyperspectral unmixing based on
  sparsity-constrained nonnegative matrix factorization with adaptive total
  variation,'' in \emph{IGARSS 2019-2019 IEEE International Geoscience and
  Remote Sensing Symposium}.\hskip 1em plus 0.5em minus 0.4em\relax IEEE, 2019,
  pp. 2139--2142.

\bibitem{shu2015multilayer}
Z.~Shu, J.~Zhou, L.~Tong, X.~Bai, and C.~Zhao, ``Multilayer manifold and
  sparsity constrainted nonnegative matrix factorization for hyperspectral
  unmixing,'' in \emph{2015 IEEE International Conference on Image Processing
  (ICIP)}.\hskip 1em plus 0.5em minus 0.4em\relax IEEE, 2015, pp. 2174--2178.

\bibitem{tong2020adaptive}
L.~Tong, J.~Zhou, B.~Qian, J.~Yu, and C.~Xiao, ``Adaptive graph regularized
  multilayer nonnegative matrix factorization for hyperspectral unmixing,''
  \emph{IEEE J. Sel. Top. Appl. Earth Observat. Remote Sens.}, vol.~13, pp.
  434--447, 2020.

\bibitem{wang2017spatial}
X.~Wang, Y.~Zhong, L.~Zhang, and Y.~Xu, ``Spatial group sparsity regularized
  nonnegative matrix factorization for hyperspectral unmixing,'' \emph{IEEE
  Trans. Geosci. Remote Sens.}, vol.~55, no.~11, pp. 6287--6304, 2017.

\bibitem{huang2019spectral}
R.~Huang, X.~Li, and L.~Zhao, ``Spectral--spatial robust nonnegative matrix
  factorization for hyperspectral unmixing,'' \emph{IEEE Trans. Geosci. Remote
  Sens.}, vol.~57, no.~10, pp. 8235--8254, 2019.

\bibitem{sreehari2016plug}
S.~Sreehari, S.~V. Venkatakrishnan, B.~Wohlberg, G.~T. Buzzard, L.~F. Drummy,
  J.~P. Simmons, and C.~A. Bouman, ``Plug-and-play priors for bright field
  electron tomography and sparse interpolation,'' \emph{IEEE Trans. on Comput.
  Imaging}, vol.~2, no.~4, pp. 408--423, 2016.

\bibitem{teodoro2018convergent}
A.~M. Teodoro, J.~M. Bioucas-Dias, and M.~A. Figueiredo, ``A convergent image
  fusion algorithm using scene-adapted gaussian-mixture-based denoising,''
  \emph{IEEE Trans. Image Process.}, vol.~28, no.~1, pp. 451--463, 2018.

\bibitem{wang2020learning}
X.~Wang, J.~Chen, C.~Richard, and D.~Brie, ``Learning spectral-spatial prior
  via 3ddncnn for hyperspectral image deconvolution,'' in \emph{Proc. IEEE Int.
  Conf. Acoust., Speech, Signal Process. (ICASSP), 2020}.\hskip 1em plus 0.5em
  minus 0.4em\relax IEEE, 2020, pp. 2403--2407.

\bibitem{Nascimento2003Vertex}
J.~M.~P. Nascimento and J.~M.~B. Dias, ``Vertex component analysis: A~fast
  algorithm to extract endmembers spectra from hyperspectral data,'' in
  \emph{Pattern Recognition and Image Analysis, First Iberian Conference,
  IbPRIA 2003, Puerto de Andratx, Mallorca, Spain, June 4-6, 2003,
  Proceedings}, 2003.

\bibitem{iordache2012total}
M.~Iordache, J.~M. Bioucas-Dias, and A.~Plaza, ``Total variation spatial
  regularization for sparse hyperspectral unmixing,'' \emph{IEEE Trans. Geosci.
  Remote Sens.}, vol.~50, no.~11, pp. 4484--4502, 2012.

\bibitem{Lina2019Regularization}
L.~Zhuang, L.~C., F.~M.~A. T., and J.~M. Bioucas-Dias, ``Regularization
  parameter selection in minimum volume hyperspectral unmixing,'' \emph{IEEE
  Trans. Geosci. Remote Sens.}, vol.~PP, no.~99, pp. 1--20, 2019.

\bibitem{heinz2001fully}
D.~C. Heinz \emph{et~al.}, ``Fully constrained least squares linear spectral
  mixture analysis method for material quantification in hyperspectral
  imagery,'' \emph{IEEE Trans. Geosci. Remote Sens.}, vol.~39, no.~3, pp.
  529--545, 2001.

\end{thebibliography}

\end{document}